\pdfoutput=1
\documentclass[runningheads]{llncs}
\usepackage{graphicx}
\usepackage{comment}
\usepackage{amsmath,amssymb} 
\usepackage{color}
\usepackage{rotating}
\usepackage{floatrow}
\usepackage{commath}
\usepackage{soul}
\usepackage{multirow,graphicx,array}
\usepackage{subcaption}
\usepackage{array, caption, floatrow, tabularx, makecell, booktabs}%
\setcellgapes{3pt}
\begin{document}
\pagestyle{headings}
\mainmatter
\def\ECCVSubNumber{720}  

\title{VPN: Learning Video-Pose Embedding for Activities of Daily Living} 

\titlerunning{VPN: Learning Video-Pose Embedding for Activities of Daily Living}
%
\author{Srijan Das \and
Saurav Sharma \and
Rui Dai \and Fran\c cois Br\'emond \and Monique Thonnat}
\authorrunning{S. Das et al.}
%
\institute{INRIA Universit\'{e} Nice C\^ote d'Azur, France\\
\email{\tt\small name.surname@inria.fr}}
\maketitle

\begin{abstract}
   In this paper, we focus on the spatio-temporal aspect of recognizing Activities of Daily Living (ADL).
   ADL have two specific properties (i) subtle spatio-temporal patterns and (ii) similar visual patterns varying with time. Therefore, ADL may look very similar and often necessitate to look at their fine-grained details to distinguish them. Because the recent spatio-temporal 3D ConvNets are too rigid to capture the subtle visual patterns across an action, we propose a novel Video-Pose Network: \textbf{VPN}. 
   The 2 key components of this VPN are a spatial embedding and an attention network. The spatial embedding projects the 3D poses and RGB cues in a common semantic space. This enables the action recognition framework to learn better spatio-temporal features exploiting both modalities. In order to discriminate similar actions, the attention network provides two functionalities - (i) an end-to-end learnable pose backbone exploiting the topology of human body, and (ii) a coupler to provide joint spatio-temporal attention weights across a video.
   Experiments\footnote{Code/models: \small{\url{https://github.com/srijandas07/VPN}}} show that VPN outperforms the state-of-the-art results for action classification on a large scale human activity dataset: \textbf{NTU-RGB+D 120}, its subset \textbf{NTU-RGB+D 60}, a real-world challenging human activity dataset: \textbf{Toyota Smarthome} and a small scale human-object interaction dataset \textbf{Northwestern UCLA}.

\end{abstract}

\section{Introduction}
Monitoring human behavior requires fine-grained understanding of actions. Activities of Daily Living (ADL) may look simple but their recognition is often more challenging than activities present in sport, movie or Youtube videos. 
 ADL often have very low inter-class variance making the task of discriminating them from one another very challenging. The challenges characterizing ADL are illustrated in fig~\ref{samples}: (i)  short and subtle actions like \textit{pouring water} and \textit{pouring grain} while \textit{making coffee} ;
(ii) actions exhibiting similar visual patterns while differing in motion patterns like \textit{rubbing hands} and \textit{clapping}; and finally, (iii) actions observed from different camera views.
In the recent literature, the main focus is the recognition of actions from internet videos~\cite{i3d,nonlocal,slow_fast,TSN,video_transformer_network} and very few studies have attempted to recognize ADL in indoor scenarios~\cite{timeception,glimpse,STA_iccv}.

\begin{minipage}{\textwidth}
  \begin{minipage}[b]{0.49\textwidth}
    \centering
    \includegraphics[width=0.95\linewidth]{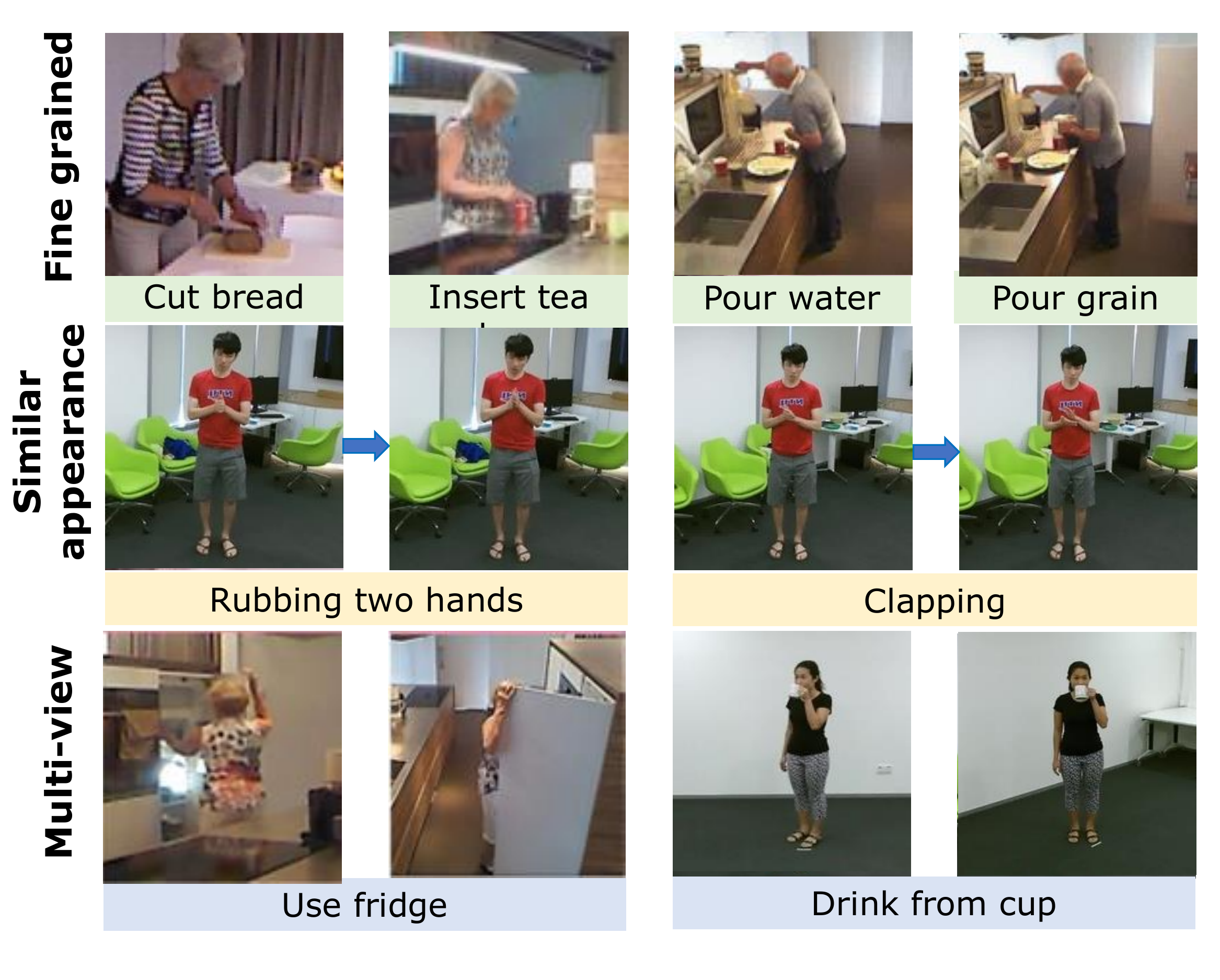}
    \captionof{figure}{Illustration of the challenges in Activities of Daily Living: fine-grained actions (top),  actions with similar visual pattern (middle) and actions viewed from different cameras (below).}
    \vspace{0.1 in}
    \label{samples}
  \end{minipage}
  \hfill
  \begin{minipage}[b]{0.49\textwidth}
    \centering
    \includegraphics[width=0.95\linewidth]{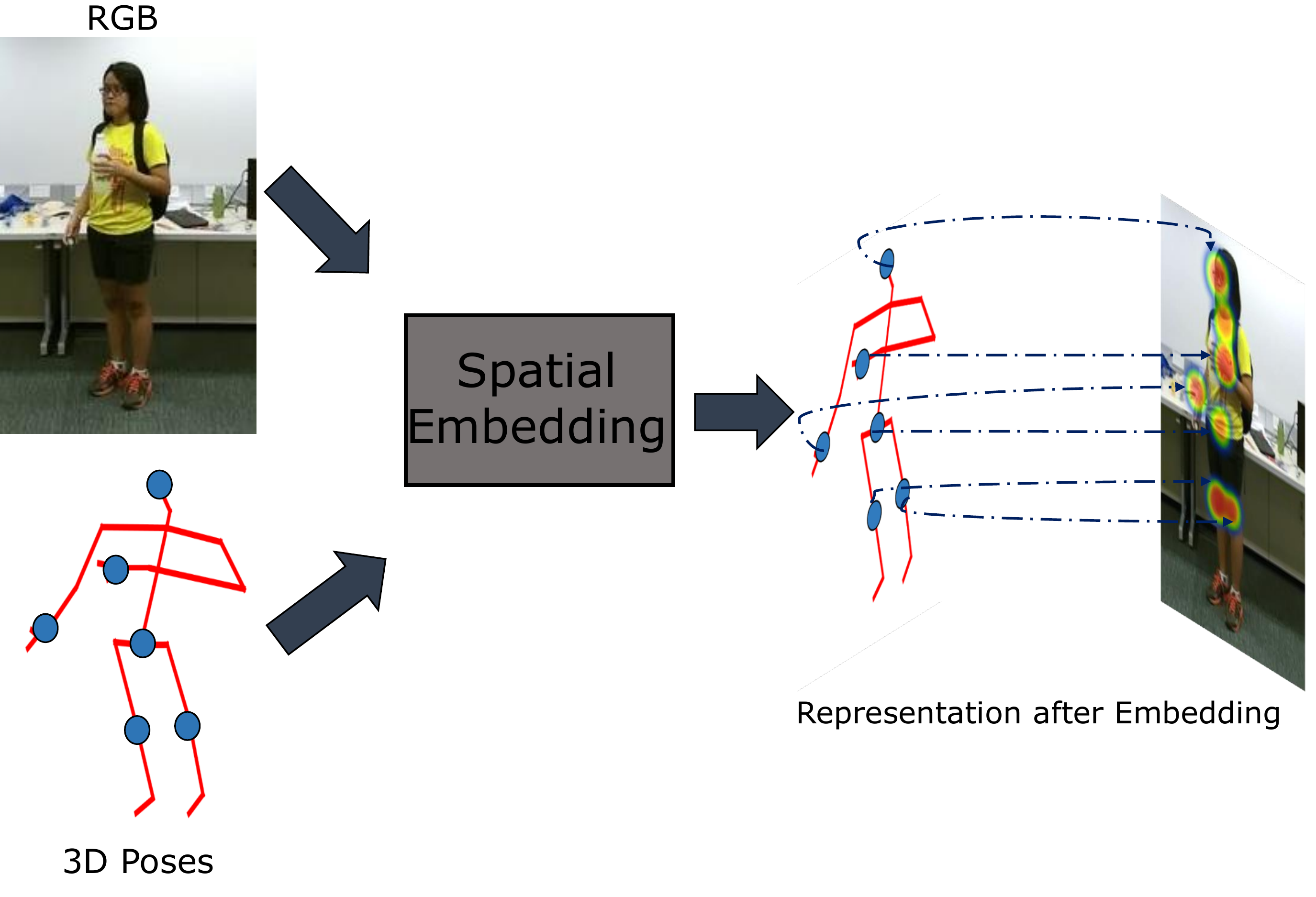}
    \captionof{figure}{Illustration of spatial embedding. Input is a RGB image and its corresponding 3D poses. For convenience, we only show 6 relevant human joints. The embedding enforces the human joints to represent the relevant regions in the image.}
    \vspace{0.1 in}
    \label{spatial_embedding}
  \end{minipage}
  \end{minipage}

For instance, state-of-the-art 3D convolutional networks like I3D~\cite{i3d} pre-trained on huge video datasets~\cite{kinetics,ucf,kuehne2011hmdb} have successfully boosted the recognition of actions from internet videos. But, these networks with similar spatio-temporal kernels applied across the whole space-time volume cannot address the complex challenges exhibited by ADL. Attention mechanisms have thus been proposed on top of these 3D convolutional networks to guide them along the regions of interest of the targeted actions~\cite{nonlocal,timeception,video_transformer_network}. 
  Following a different direction, action recognition for ADL has been dominated by the use of human 3D poses~\cite{gemetricfeaturesWACV2017,valstm}.
  They provide a strong clue for understanding the visual patterns of an action over time. 3D poses are robust to illumination changes, view adaptive and provide critical geometric information about human actions. However, they lack incorporating the appearance information which is an essential property in ADL (especially for human-object interaction).
  
  Consequently, attempts have been made to utilize 3D poses to weight the discriminative parts of a RGB feature map~\cite{Baradel_BMVC,glimpse,STA-hands,spatial-i3d,STA_iccv}. These methods have improved the action recognition performance but they do not take into account the alignment of the RGB cues and the corresponding 3D poses. Therefore, we propose a spatial embedding to project the visual features and the 3D poses in the same referential. Before describing our contribution, we answer two intuitive questions below.

\noindent\textbf{First, why is spatial embedding important?} - Previous pose driven attention networks can be perceived as guiding networks to help the RGB cues focus on the salient information for action classification. For these guiding networks, it is important to have an accurate correspondences between the poses and RGB data. So, the objective of the spatial embedding is to find correspondences between the 3D human joints and the image regions representing these joints as illustrated in fig~\ref{spatial_embedding}. This task of finding correlation between both modalities can (i) provide informative pose aware feedback to the RGB cues, and (ii) improve the functionalities of the guiding network. \\
\textbf{Second, why not performing temporal embedding?} - We argue that the need of embedding is to provide proper alignment between the modalities. Across time, the 3D poses are already aligned assuming that there is a 3D pose for every images. 
However, even if the number of 3D poses does not correspond to the number of image frames (as in~\cite{Baradel_BMVC,glimpse,STA-hands,spatial-i3d,STA_iccv}), the fact that variance in poses for few consecutive frames is negligible, especially for ADL, implies temporal embedding is not needed. 


 We propose a recognition model based on a Video-Pose Network, \textbf{VPN} to recognize a large variety of human actions. VPN consists of a spatial embedding and an attention network.
VPN exhibits the following novelties:
(i) a spatial embedding learns an accurate video-pose embedding to enforce the relationships between the visual content and 3D poses, (ii) an attention network learns the attention weights with a tight spatio-temporal coupling
for  better modulating the RGB feature map, (iii) the attention network takes the spatial layout of the human body into account by processing the 3D poses through Graph Convolutional Networks (GCNs). \\
The proposed recognition model is end-to-end trainable and our proposed VPN can be used as a layer on top of any 3D ConvNets.

\section{Related Work}
Below, we discuss the relevant action recognition algorithms w.r.t. their input modalities. \\
\textbf{RGB} - Traditionally, image level features~\cite{DT,IDT} have been aggregated over time using encoding techniques like Fisher Vector~\cite{fischer} and NetVLAD~\cite{NetVLAD}. But these video descriptors do not encode long-range temporal information. Then, temporal patterns of actions have been modelled in videos using sequential networks. These sequential networks like LSTMs are fed with convolutional features from images~\cite{lrcn} and thus, they model the temporal information based on the evolution of appearance of the human actions. However, these methods first process the image level features and then capture their temporal evolution preventing the computation of joint spatio-temporal patterns over time. 

 Due to this reason, Du et al.~\cite{C3D} have proposed 3D convolution to model the spatio-temporal patterns within an action. The 3D kernels provide tight coupling of space and time towards better action classification. 
Later on, holistic methods like I3D~\cite{i3d}, slow-fast network~\cite{slow_fast}, MARS~\cite{mars} and two-in-one stream network~\cite{dance_with_flow} have been fabricated for generic datasets like Kinetics~\cite{kinetics} and UCF-101~\cite{ucf}.
But these networks are trained globally over the whole 3D volume of a video and thus, are too rigid to capture salient features for subtle spatio-temporal patterns for ADL. 

 Recently several attention mechanisms have been proposed on top of the aforementioned 3D ConvNets to extract salient spatio-temporal patterns. For instance, Wang et al.~\cite{nonlocal} have proposed a non-local module on top of I3D which computes the attention of each pixel as a weighted sum of the features of all pixels in the space-time volume. But this module relies too much on the appearance of the actions, i.e., pixel position within the space-time volume. As a consequence, this module though effective for the classification of actions in internet videos, fails to disambiguate ADL with similar motion and fails to address view invariant challenges. \\
\textbf{3D Poses} - To focus on the view-invariant challenge, temporal evolution of 3D poses have been leveraged through sequential networks like LSTM and GRU for skeleton based action recognition~\cite{gemetricfeaturesWACV2017,st-lstm,valstm}. Taking a step ahead, LSTMs have also been used for spatial and temporal attention mechanisms to focus on the salient human joints and key temporal frames~\cite{sta_lstm}. Another framework represents 3D poses as pseudo image to leverage the successful image classification CNNs for action classification~\cite{skel_CNN_1,skel_CNN_2}. Recently, graph-based methods model the data as a graph with joints as vertexes and bones as edges~\cite{stgcn2018aaai,deep-progressive,directed_graph}. Compared to sequential networks and pseudo image based methods, graph-based methods make use of the spatial configuration of the human body joints and thus, are more effective. However, the skeleton based action recognition lacks in encoding the appearance information which is critical for ADL recognition. \\
\textbf{RGB + 3D Poses} - In order to make use of the pros of both modalities, i.e. RGB and 3D Poses, it is desirable to fuse these multi-modal information into an integrated set of discriminative features. As these modalities are heterogeneous, they must be processed by different kinds of network to show their effectiveness. This limits their performance in simple multi-modal fusion strategy~\cite{rgb+pose_1,rgb+pose_2,Luo_2018_ECCV}. As a consequence, many pose driven attention mechanisms have been proposed to guide the RGB cues for action recognition. In~\cite{STA-hands,Baradel_BMVC,glimpse}, the pose driven attention networks implemented through LSTMs, focus on the salient image features and the key frames. Then, with the success of 3D CNNs, 3D poses have been exploited to compute the attention weights of a spatio-temporal feature map.
Das et al.~\cite{spatial-i3d} have proposed a spatial attention mechanism on top of 3D ConvNets to weight the pertinent human body parts relevant for an action. Then, authors in~\cite{STA_iccv} have proposed a more general spatial and temporal attention mechanism in a dissociated manner. 
But these methods have the following drawbacks: 
(i) there is no accurate correspondence between the 3D poses and the RGB cues in the process of computing the attention weights~\cite{STA-hands,Baradel_BMVC,glimpse,spatial-i3d,STA_iccv};
(ii) the attention sub-networks~\cite{STA-hands,Baradel_BMVC,glimpse,spatial-i3d,STA_iccv} neglect the topology of the human body while computing the attention weights;
(iii) the attention weights in~\cite{spatial-i3d,STA_iccv} provide identical spatial attention along the video. As a result, action pairs with similar appearance like \textit{jumping} and \textit{hopping} are mis-classified.

In contrast, we propose a new spatial embedding to enforce the correspondences between RGB and 3D pose which has been missing in the state-of-the-art methods. The embedding is built upon an end-to-end learnable attention network. The attention network considers the human topology to better activate the relevant body joints for computing the attention weights. To the best of our knowledge, none of the previous action recognition methods have combined human topology with RGB cues. In addition, the proposed attention network couples the spatial and temporal attention weights in order to provide spatial attention weights varying along time.

\begin{figure*}
\centering
\includegraphics[width=1\linewidth]{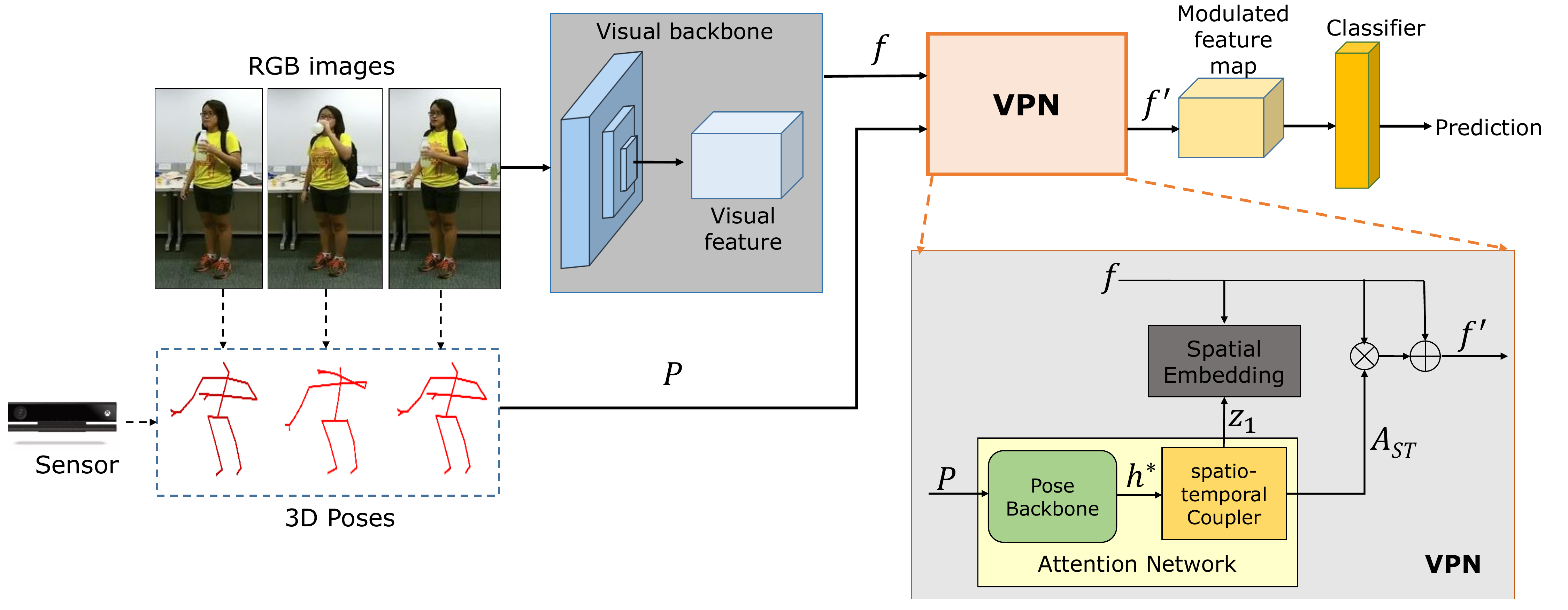}
\caption{\textbf{Proposed Action Recognition Model}: Our model takes as input RGB images with their corresponding 3D poses. The RGB images are processed by a visual backbone which generates a spatio-temporal feature map ($f$). The proposed \textbf{VPN} takes as input the feature map ($f$) and the 3D poses ($P$). VPN consists of two components: an attention network and a spatial embedding. The attention network further consists of a Pose Backbone and a spatio-temporal Coupler. VPN computes a modulated feature map $f'$. This modulated feature map $f'$ is then used for classification. }\vspace{-0.1em}
\label{VPN}
\end{figure*}

\section{Proposed Action Recognition Model}
Our objective is to design an accurate spatial embedding of poses and visual content to better extract the discriminative spatio-temporal patterns. As shown in fig.~\ref{VPN}, the input of our proposed recognition model are the RGB images and their 3D poses. The 3D poses are either extracted from depth sensor or from RGB using LCRNet~\cite{lcrnet_new}.  The proposed Video-Pose Network \textbf{VPN} takes as input the visual feature map and the 3D poses. Below, we discuss the action recognition model in details.
\subsection{Video Representation}
Taking as input a stack of human cropped images from a video clip, the spatio-temporal representation $f$ is computed by a 3D convolutional network (the visual backbone in fig.~\ref{VPN}). $f$ is a feature map of dimension $t_c \times m \times n \times c$, where $t_c$ denotes the temporal dimension, $m \times n$ the spatial scale and $c$ the channels. Then, the feature map $f$ and the corresponding poses $P$ are processed by the proposed network.

\subsection{VPN}
\begin{figure*}
\centering
\includegraphics[width=1\linewidth]{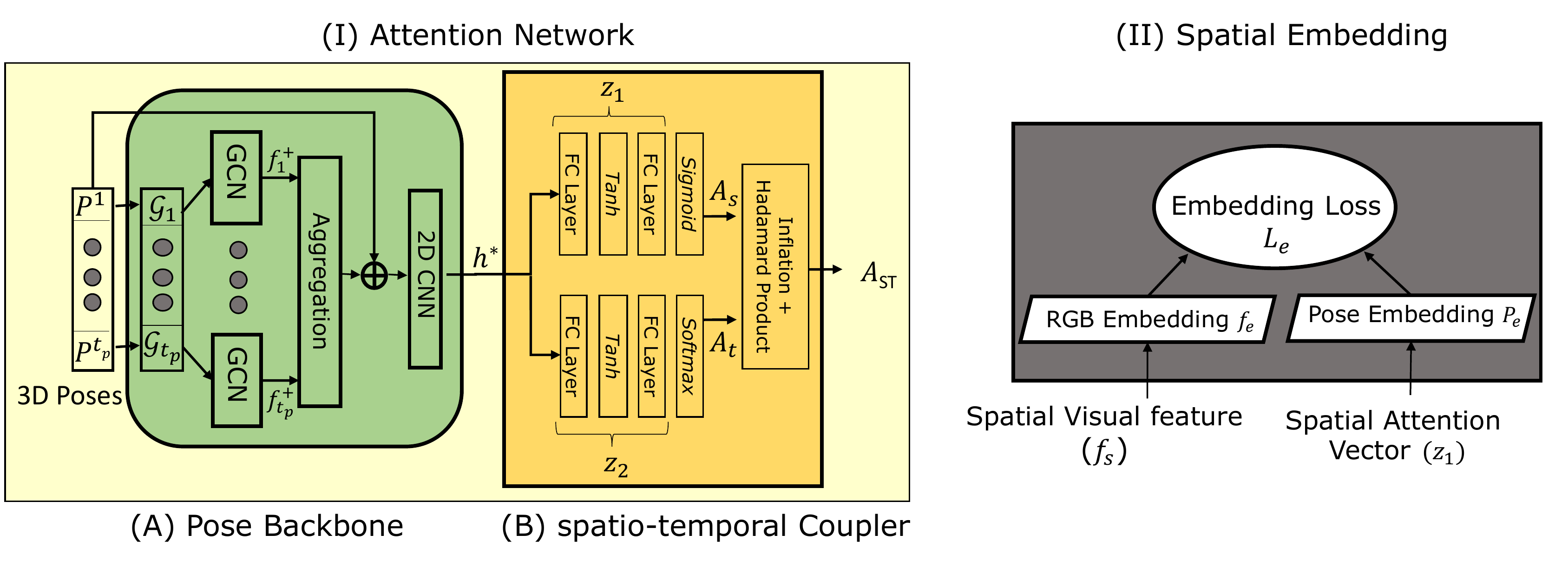}
\caption{The components in VPN: (I) Attention Network (left) and (II) Spatial Embedding (right). We present a zoom of the attention Network with: (A) a GCN Pose Backbone, and (B) a spatio-temporal Coupler to generate spatio-temporal attention weights $A_{ST}$ }\vspace{-1em}
\label{VPN_modules}
\end{figure*}

VPN can be thought as a layer which can be placed on top of any 3D convolutional backbone. VPN takes as input a 3D feature map ($f$) and its corresponding 3D poses ($P$) to perform two functionalities. First, to provide an accurate alignment of the human joints with the feature map $f$. Second, to compute a modulated feature map ($f'$) which is further classified for action recognition. The modulated feature map ($f'$) is weighted along space and time as per its relevance. VPN exploits the highly informative 3D pose information to transform the visual feature map $f$ and finally, compute the attention weights. This network has two major components as shown in fig~\ref{VPN_modules}: (I) an attention network and (II) a spatial embedding.
Though the intrinsic parameters of the attention network and the spatial embedding learns in parallel, we present these two components in the following order for better understanding.


\subsubsection{(I) Attention Network -}
The attention network consists of a Pose Backbone and a spatio-temporal Coupler. Such a framework for pose driven attention network is unique compared to the other state-of-the-art methods using poses and RGB. The proposed attention network unlike~\cite{Baradel_BMVC,glimpse,spatial-i3d,STA_iccv} takes into account the human spatial configuration and it also learns coupled spatio-temporal attention weights for the visual feature map $f$. 
\subsubsection{Pose Backbone -}
The input poses along the video are processed in a Pose Backbone.
The pose based input of VPN are the 3D human joint coordinates $P \in \mathbb{R}^{3\times J \times t_p}$ stacked along $t_p$ temporal dimension, where $J$ is the number of skeleton joints. The Pose Backbone processes these 3D poses to compute pose features $h^*$ which are used further in the attention network for computing the spatio-temporal attention weights. 
They carry meaningful information in a compact way, so the proposed attention network can efficiently focus on salient action parts.


 For the Pose Backbone, we use \textbf{GCNs} to learn the spatial relationships between the 3D human joints to provide attention weights to the visual feature map ($f$). We aim at exploiting the graphical structure of the 3D poses. 
In fig.~\ref{VPN_modules}(I), we illustrate our GCN pose backbone (marked (A)). 
For each pose input $P_t \in \mathbb{R}^{3\times J}$ with $J$ joints, we first construct a graph  $\mathcal{G}_{t}(P_t, E)$ where $E$ is the $J \times J$ weighted adjacency matrix:
\[
    e_{ij}= 
\begin{cases}
    0,              & \text{if } i=j\\
    \alpha,              & \text{if joint i and joint j are connected} \\
    \beta,              & \text{if joint i and joint j are disconnected} \\
\end{cases}
\]
Each graph $\mathcal{G}_{t}$ at time $t$ is processed by a GCN to compute feature $f^{+}_t$: 
\begin{equation}
    f^{+}_t = D^{-\frac{1}{2}}(E + I)D^{-\frac{1}{2}}\mathcal{G}_{t}W_t,
\end{equation}
where $W_t$ is the weight matrix and $D$ is the diagonal degree matrix with $D_{ii} = \Sigma_j(E_{ij} + I_{ij})$ its diagonal elements. For all $t=1, 2, ..., t_p$, the GCN output features $f^{+}_t$ are aggregated along time, resulting in a 3D tensor $[f^{+}_1, f^{+}_2, ..., f^{+}_{t_p}]$. 
\noindent Finally, the 3D pose tensor is combined with the original pose input by a residual connection followed by a set of convolutional operations. 
Now, the GCN pose backbone provides salient features $h^*$ because of its use of the graphical structure of the 3D joints.
\subsubsection{Spatio-temporal Coupler -}
 The attention network in VPN learns the spatio-temporal attention weights from the output of Pose Backbone in two steps as shown in fig.~\ref{VPN_modules}(I)(B). In the first step, the spatial and temporal attention weights ($A_S$ and $A_T$) are classically trained as in~\cite{sta_lstm} to get the most important body part and key frames for an action.
 The output feature $h^*$ of Pose Backbone follows two separate non-linear mapping functions to compute the spatial and temporal attention weights. These spatial $A_S$ and temporal $A_T$ weights are defined as 
\begin{equation}
    A_S = \sigma(z_1); \hspace{0.25in} A_T = softmax(z_2)
\end{equation}
where $z_r = W_{z_r}tanh(W_{h_r}h^* + b_{h_r}) + b_{z_r}$ (for $r = 1,2$) with subscripted $W$ and $b$, the corresponding weights and biases are the latent spatial and temporal attention vectors.
The dissociated attention weights $A_S$ and $A_T$ having dimension $m \times n$ and $t_c$ respectively, can undergo a linear mapping to obtain spatially and temporally modulated feature maps. The resultant model is equivalent to the separable STA model~\cite{STA_iccv}. In contrast, we propose to further perform a coupling of the spatial and temporal attention weights. 
Thus in the second step, joint spatio-temporal attention weights are computed by performing a Hadamard product on the spatial and temporal attention weights. In order to perform this matrix multiplication, the spatial and temporal attention weights are inflated by duplicating the same attention weights in temporal and spatial dimension respectively. Hence, the $m \times n \times t_c$ dimensional spatio-temporal attention weights $A_{ST}$ are obtained by $A_{ST}=inflate(A_S) {\circ} inflate(A_T)$. This two-step attention learning process enables the attention network to compute spatio-temporal attention weights in which the spatial saliency varies with time. The obtained attention weights are crucial to disambiguate actions with similar appearance as they may have dissimilar motion over time. 

 Finally, the spatio-temporal attention weights $A_{ST}$ are linearly multiplied with the input video feature map $f$, followed by a residual connection with the original feature map $f$ to output the modulated feature map $f'$. The residual connection enables the network to retain the properties of the original visual features.
\subsubsection{(II) Spatial Embedding of RGB and Pose - }
The objective of the embedding model is to provide tight correspondences between both pose and RGB modalities used in VPN. The state-of-the-art methods~\cite{spatial-i3d,STA_iccv} attempt to provide the attention weights on the RGB feature map using 3D pose information without projecting them into the same 3D referential. The mapping with the pose is only done by cropping the person within the input RGB images.
The spatial attention computed through the 3D joint coordinates does not correspond to the part of the image (no pixel to pixel correspondence), although it is crucial for recognizing fine-grained actions. To correlate both modalities, an embedding technique inspired from image captioning task~\cite{text_video_embedding,joint_modelling_captioning} is used to build an accurate RGB-Pose embedding in order to enable the poses to represent the visual content of the actions (see fig.~\ref{VPN_modules}(II)).

 We assume that a low dimensional embedding exists for the global spatial representation of video feature map $f_s = \Sigma_{i=1}^{t_c}f(i,:,:,:)$ (a $D_v$ dimensional vector) and its corresponding pose based latent spatial attention vector $z_1$ (a $D_p$ dimensional vector). The mapping function can be derived from this embedding by
\begin{equation}
    f_e = T_vf_s    \hspace{0.1in}   and \hspace{0.1in} P_e = T_pz_1, 
\end{equation}
where $T_v \in R^{D_e \times D_v}$ and $T_p \in R^{D_e \times D_p}$ are the transformation matrices that project the video content and the 3D poses into the common $D_e$ dimensional embedding space. This mapping function is applied on the global spatial representation of the visual feature map and the pose based features in order to attain the aforementioned objective of the spatial embedding. 

\noindent To measure the correspondence between the video content and the 3D poses, we compute the distance between their mappings in the embedding space. Thus, we define an embedding loss as a hypersphere feature metric space 
\begin{equation}
    L_e = ||\widehat{T_vf_s} - \widehat{T_pz_1}||_2^2   \hspace{0.4in} s.t. \hspace{0.1in} ||T_v||_2 = ||T_p||_2 = 1
\end{equation}
$\widehat{T_vf_s} = \frac{T_vf_s}{||T_vf_s||_2}$ and $\widehat{T_pz_1} = \frac{T_pz_1}{||T_pz_1||_2}$ are the feature representations projected to the unit hypersphere. The norm constraint $||T_v||_2 = 1 $ \& $||T_p||_2 = 1$ simply prevents the trivial solution $\hat{T_v} = \hat{T_p} = 0$. This embedding loss along with the global classification loss provides a linear transformation on the RGB feature map that preserves the low-rank structure for the action representation and introduces a maximally separated features for different actions. Now, the kernels at the visual backbone are updated with a gradient proportional to $(f_e - P_e)$, which in turn transforms the visual feature map to learn pose aware characteristics.
Consequently, we strengthen the correspondences between video and poses by minimizing the embedding loss. This embedding ensures that the pose information to be used for computing the spatial attention weights aligns with the content of the video.

 Note that the embedding loss also provides feedback to the pose based latent spatial attention vectors ($z_1$), which in turn transfers knowledge from the 2D image space to pose 3D referential. This allows the attention network to provide better and meaningful spatial attention weights ($A_s$) compared to the attention network without the embedding. We will quantify this observation in the experiments.

\subsection{Training jointly the 3D ConvNet and VPN}
VPN can be trained as a layer on top of any 3D ConvNet. The 3D ConvNet can be pre-trained for the action classification task for faster convergence. Finally, VPN is plugged into the 3D ConvNet for an end-to-end training with a regularized loss $L$ formulated as
\begin{equation}
    L = \lambda_1L_C + (1-\lambda_1)L_e + \lambda{_{2}}L_a
\end{equation}
Here, $L_C$ is the cross-entropy loss, $L_e$ is the embedding loss; the trade-off between these two losses is captured by linear fusion with a positive parameter $\lambda_1$;
$L_a$ is the attention regularizer with $\lambda_2$ weighting factor. The attention regularizer consists of the spatial and temporal attention weight regularizer and is formulated as 
\begin{equation}
    L_a = \sum_{j=1}^{m \times n}\norm{A_{s}(j)}_{2} + \sum_{j=1}^{t_c}(1-A_{t_c}(j))^2
\end{equation}
This additional regularization term $L_a$ ensures that the attention weights are not biased to provide extremely high values to the parts of the spatio-temporal feature map with more relevance and completely neglecting the other parts. \\

\section{Experiments}
We evaluate the effectiveness of our model for action classification. We consider four public datasets which are the popular datasets for ADL: NTU-60~\cite{NTU_RGB+D}, NTU-120~\cite{ntu120}, Toyota-Smarthome~\cite{STA_iccv} and Northwestern-UCLA~\cite{nucla}.

\noindent\textbf{NTU RGB+D} (NTU-60 \& NTU-120): NTU-60 is acquired with a Kinect v2 camera and consists of 56880 video samples with 60 activity classes. The activities were performed by 40 subjects and recorded from 80 viewpoints. For each frame, the dataset provides RGB, depth and a 25-joint skeleton of each subject in the frame. For evaluation, we follow the two protocols proposed in~\cite{NTU_RGB+D}: cross-subject (CS) and cross-view (CV). NTU-120 is a super-set of NTU-60 adding a lot of new similar actions. NTU-120 dataset contains 114k video clips of 106 distinct subjects performing 120 actions in a laboratory environment with 155 camera views. For evaluation, we follow a cross-subject ($CS_1$) protocol and a cross-setting ($CS_2$) protocol proposed in~\cite{ntu120}.

\noindent\textbf{Toyota-Smarthome} (Smarthome) is a recent ADL dataset recorded in an apartment where 18 older subjects carry out tasks of daily living during a day. The dataset contains 16.1k video clips, 7 different camera views and 31 complex activities performed in a natural way without strong prior instructions. This dataset provides RGB data and 3D skeletons which are extracted from LCRNet~\cite{lcrnet_new}. For evaluation
 on this dataset, we follow cross-subject ($CS$) and cross-view ($CV_1$ and $CV_2$) protocols proposed in~\cite{STA_iccv}.

\noindent\textbf{Northwestern-UCLA Multiview activity 3D Dataset} (N-UCLA) is acquired simultaneously by three Kinect v1 cameras. The dataset consists of 1194 video samples with 10 activity classes. The activities were performed by 10 subjects, and recorded from three viewpoints. We performed experiments on N-UCLA using the cross-view (CV) protocol proposed in~\cite{nucla}: we trained our model on samples from two camera views and tested on the samples from the remaining view. For instance, the notation $V_{1,2}^3$ indicates that we trained on samples from view 1 and 2, and tested on samples from view 3. \\
\noindent The presence of ADL challenges like fine-grained and similar appearance activities is in higher magnitude in NTU-120 and Smarthome datasets. So, we perform all our ablation studies on these two datasets. We abbreviate Smarthome as SH in table~\ref{ablation},~\ref{ablation_gcn},~\ref{ablation_loss} and~\ref{ablation_SA}.

\subsection{Implementation details}
\textbf{Training.} In our experiments, the selected \textbf{visual backbone} is I3D~\cite{i3d} network pre-trained on ImageNet~\cite{imagenet_cvpr09} and Kinetics-400~\cite{kinetics}. The visual backbone takes 64 video frames as input. The input of the \textbf{VPN} consists of the feature map extracted from \texttt{Mixed\_5c} layer of I3D and the corresponding 3D poses.

\noindent The \textbf{pose backbone} takes as input a sequence of $t_p$ 3D poses uniformly sampled from each clip. Hyper-parameter $t_p = 20, 20, 30$ and $5$ for NTU-60, NTU-120, Smarthome and N-UCLA respectively. 
\noindent For the pose backbone, we use $t_p$ number of GCNs, each processing a pose from the sequence. The weighting parameters $\alpha$ and $\beta$ for computing the adjacency matrix of the pose based graph are set to 5 and 2 respectively. GCN projects the input joint coordinates to a $64-dimensional$ space. The output of the GCN is passed to a set of convolutional operations (see fig.~\ref{VPN_modules}(I)(A)) which consists of three 2D convolutional layers each are followed by a Batch Normalization layer and a ReLU layer. The output channels of the convolutional layers are 64, 64 and 128. 

\noindent For classification, a global-average pooling layer followed by a dropout~\cite{Dropout} of 0.3 and a \textit{softmax} layer are added at the end of the recognition model for class prediction. Our recognition model is trained with a 4-GPU machine where each GPU has 4 video clips in a mini-batch. Our model is trained for 30 epochs in total, with SGD optimizer having initial learning rate of 0.01 and decay rate of 0.1 after every 10 epochs. The trade off ($\lambda_1$) and regularizer ($\lambda_2$) parameters are set to 0.8 and 0.00001 respectively for all the experiments.


\noindent \textbf{Inference.} For the recognition model, we perform fully convolutional inference in space as in~\cite{nonlocal}. The final classification is obtained by max-pooling the softmax scores.

\begin{table}[htp]
\floatsetup{floatrowsep=qquad,captionskip=5pt} \tabcolsep=4pt
\begin{floatrow}
\floatsetup{floatrowsep=qquad,captionskip=5pt} \tabcolsep=4pt
\begin{floatrow}
\ttabbox{\caption{Ablation study to show the effectiveness of each VPN component.}}{%
\scalebox{0.65}{
\begin{tabular}{l|ccccc}
\hline
 VPN components    & NTU-120  & NTU-120 & \small{SH} & \small{SH} \\
              &     $CS_1$    &   $CS_2$     & CS & $CV_2$  \\
\hline
$l_1$: visual backbone   & 77.0 & 80.1 & 53.4 & 45.1 \\
\hline
$l_2$: $l_1$ + attention network  & 85.4 & 86.9  & 56.4 & 50.5 \\
\hline
$l_3$: $l_2$ + spatial embedding   &   \textbf{86.3} & \textbf{87.8} & \textbf{60.8} & \textbf{53.5} \\
\hline
\end{tabular}
}
\label{ablation}
}
\end{floatrow}

\hspace{4pt}
\floatsetup{floatrowsep=qquad,captionskip=5pt} \tabcolsep=4pt

\begin{floatrow}
\ttabbox{\caption{Performance of VPN with different choices of Attention Network.}}{%
\scalebox{0.58}{
\begin{tabular}{l|c|c|ccccc}
\hline
Model    & Pose & Coupler &NTU-120  & NTU-120 & \small{SH} & \small{SH} \\
        & Backbone &     &     $CS_1$    &   $CS_2$     & CS & $CV_2$  \\
\hline
$l_4$: VPN   & LSTM & $\times$ & 84.7 & 83.6 & 57.1 & 50.6  \\
\hline
$l_5$: VPN & GCN & $\times$ & 85.6 & 86.8  & 60.1 & 53.1 \\
\hline
$l_6$: VPN   & LSTM & \checkmark & 85.3 & 84.1 & 57.6 & 51.5  \\
\hline
$l_7$: VPN   & GCN & \checkmark &   \textbf{86.3} & \textbf{87.8} & \textbf{60.8} & \textbf{53.5} \\
\hline
\end{tabular}
}
\label{ablation_gcn}
}
\end{floatrow}

\end{floatrow}
\end{table}
\begin{table}[htp]
\floatsetup{floatrowsep=qquad,captionskip=5pt} \tabcolsep=4pt
\begin{floatrow}
\floatsetup{floatrowsep=qquad,captionskip=5pt} \tabcolsep=4pt
\begin{floatrow}
\ttabbox{\caption{Performance of VPN with different embedding losses $l_e$.}}{%
\scalebox{0.62}{
\begin{tabular}{l|ccccc}
\hline
Loss    & NTU-120  & NTU-120 & \small{SH} & \small{SH} \\
              &     $CS_1$    &   $CS_2$     & CS & $CV_2$  \\
\hline
KL-divergence $D_{KL}(f_e||P_e)$  & 85.5 & 87.1 & 57.2 & 50.9 \\
KL-divergence $D_{KL}(P_e||f_e)$  & 85.6 & 86.9 & 57.0 & 51.1 \\
\hline
Bi-directional KL-divergence   & 86.1 & 87.2 & 57.2 & 51.7 \\
Normalized Euclidean loss & \textbf{86.3} & \textbf{87.8} & \textbf{60.8} & \textbf{53.5} \\
\hline
\end{tabular}
}
\label{ablation_loss}
}
\end{floatrow}

\hspace{4pt}
\floatsetup{floatrowsep=qquad,captionskip=5pt} \tabcolsep=4pt

\begin{floatrow}
\ttabbox{\caption{Impact of Spatial Embedding on Spatial Attention.}}{%
\scalebox{0.58}{
\begin{tabular}{l|c|c|ccccc}
\hline
Model    & Pose & Spatial &NTU-120  & NTU-120 & \small{SH} & \small{SH} \\
        & Backbone & Embedding     &     $CS_1$    &   $CS_2$     & CS & $CV_2$  \\
\hline
VPN   & LSTM & $\times$ & 81.7 & 81.2 & 45.5 & 50.0  \\
VPN   & LSTM & \checkmark & 82.7 & 82.0 & 56.5 & 52.6  \\
\hline
VPN & GCN & $\times$ & 82.6 & 84.3  & 49.1 & 51.7 \\
VPN & GCN & \checkmark & \textbf{83.1} & \textbf{85.3} & \textbf{58.4} & \textbf{53.1} \\
\hline

\hline
\end{tabular}
}
\label{ablation_SA}
}
\end{floatrow}

\end{floatrow}
\end{table}

\subsection{Ablation Study}
\vspace{-0.1 in}
Our model includes two novel components, the spatial embedding and the attention network. Both of them are critical for good performance on ADL recognition. We show the importance of the attention network and the spatial embedding of VPN in table~\ref{ablation}. We also show the effectiveness of the spatial embedding with different instantiation of the attention network in table~\ref{ablation_gcn}.  


\noindent \textbf{How effective is VPN?}
In order to answer this point, we show the action classification accuracy with baseline I3D ($l_1$) which is the visual backbone and then incorporate the VPN components: the attention network ($l_2$) and the spatial embedding ($l_3$) one-by-one in table~\ref{ablation}. The attention network ($l_2$) improves significantly the classification of the actions (upto 8.4\% on NTU-120 and 5.4\% on Smarthome) by providing spatio-temporal saliency to the I3D feature maps. With the spatial embedding ($l_3$), the action classification further improves (upto 0.9\% on NTU-120 and 4.4\% on Smarthome).


\noindent \textbf{Diagnosis of the attention network -}
In table~\ref{ablation_gcn}, we further illustrate the importance of each component in the attention network, i.e. the Pose Backbone and the spatio-temporal coupler.
We have designed a baseline attention network with LSTM as pose backbone following~\cite{STA_iccv}. We compare the LSTM pose backbone in $l_4$ and $l_6$ with our proposed GCN instantiation in $l_5$ and $l_7$. The attention network without a spatio-temporal coupler provides dissociated spatial and temporal attention weights in $l_4$ and $l_5$ in contrast to our proposed coupler in $l_6$ and $l_7$.
Firstly, we observe that the GCN pose backbone makes use of the human joint topology, thus improves the classification accuracy in all scenarios with or without the coupler. Consequently, actions like \textit{Snapping Finger} (+24.5\%) and \textit{Apply cream on face} (+23.9\%) improves significantly with GCN instantiation ($l_6$) compared to LSTM ($l_7$).
Secondly, we observe that the spatio-temporal coupler provides fine spatial attention weights for the most important frames in a video, which enables the model to disambiguate actions with similar appearance but dissimilar motion. Consequently, the coupler ($l_7$) improves the classification accuracy up to 1\% on NTU-120 and 0.7\% on Smarthome w.r.t. dissociating the attention weights ($l_5$). 
For instance, with dissociation of the attention weights, \textit{rubbing two hands} was confused with \textit{clapping} and \textit{flicking hair} was confused with \textit{putting on headphone}.
With VPN, the coupler improves the classification accuracy of actions \textit{rubbing two hands} and \textit{flicking hair} by 25\% and 19.6\% respectively. 

\noindent \textbf{Which loss is better for learning the spatial embedding?} In this ablation study (Table~\ref{ablation_loss}), we compare different losses for projecting the 3D poses and RGB cues in a common semantic space. First, we compare the KL-divergence losses~\cite{uni_KL_1,uni_KL_2} ($D_{KL}(f_e||P_e)$ and $D_{KL}(P_e||f_e)$) from $P_e$ to $f_e$ and vice-versa. Then, we compare a bi-directional KL-divergence loss~\cite{bi_KL_1,bi_KL_2,bi_KL_3} ($D_{KL}(f_e||P_e)$ + $D_{KL}(P_e||f_e)$) to our normalized euclidean loss. We observe that (i) the loss using $D_{KL}(f_e||P_e)$ and $D_{KL}(P_e||f_e)$ deteriorates the action classification accuracy as the feedback is in one direction either towards RGB or poses, implying two-way feedback for the visual features and the attention network is necessary, (ii) our normalized euclidean loss outperforms the bi-directional KL divergence loss, exhibit its superiority. 

\noindent \textbf{Impact of Embedding on Spatial attention -} In table~\ref{ablation_SA}, we show the impact of spatial embedding on the attention network providing spatial attention only. We perform the experiments with different choice of Pose Backbone, i.e. LSTM as discussed above and our proposed GCN. The spatial embedding provides a tight correspondence between the RGB data and poses. As a result, it boosts the classification accuracy in all the experiments. It is worth noting that the improvement is significant for Smarthome as it contains many fine-grained actions with videos captured by fixed cameras in an unconstrained Field of View. Thus, enforcing the embedding loss enhances the spatial precision during inference. As a result, the classification accuracy of fine-grained actions like \textit{pouring water} (+77.7\%), \textit{pouring grains} (+76.1\%) for making coffee, \textit{cutting bread} (+50\%), \textit{pouring from kettle} (+42.8\%) and \textit{inserting teabag} (+35\%) improves VPN with GCN pose backbone compared to its counterpart without embedding.

\begin{figure}
\centering
\includegraphics[width=1\linewidth]{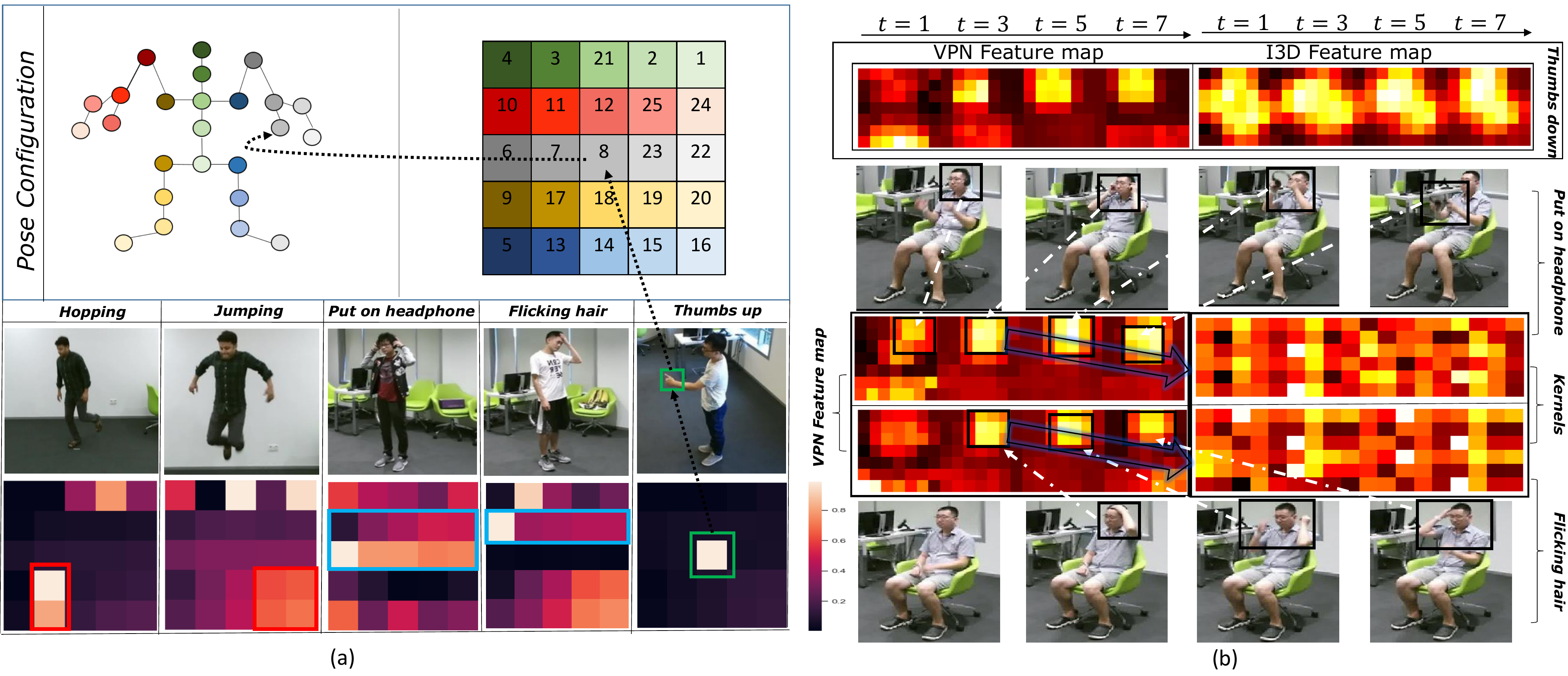}
\caption{ (a) The heatmaps of the activations of the 3D joint coordinates (output of GCN) in the attention network of VPN. The area in the colored bounding boxes shows that different joints are activated for similar actions. (b) Heatmaps of visual feature maps \& corresponding activated kernels for different time stamps. These heatmaps show that VPN has better discriminative power than I3D.}\vspace{-1em}
\label{heatmap}
\end{figure}
\subsection{Qualitative Analysis}
Fig.~\ref{heatmap}(a) visualizes the activation of the human joints at the output of pose backbone (with GCNs) in VPN. The figure depicts the activations of the 3D joints. They are presented in a sequence of the human body topological order (follow first row of fig.~\ref{heatmap}(a)) for convenient visualization. VPN is able to disambiguate actions with similar appearance like \textit{hopping} and \textit{jumping} due to high order activation at relevant joints of the human legs. The discriminative leg joints with high activation have been marked with a red bounding box in fig.~\ref{heatmap}(a) (third row). Similarly, for actions like \textit{put on headphone} with two hands and \textit{flicking hair} with one hand, the blue bounding boxes demonstrate high activation of both the hand joints for the former action as compared to high activation of a single hand joints for the latter. For a very fine-grained action like \textit{thumbs up}, the thumb joint is highly activated as compared to the other joints. This shows that the GCN pose backbone in VPN is a crucial ingredient for better action recognition. \\
\noindent In fig.~\ref{heatmap}(b), we compare the heatmap of the VPN and I3D feature maps for different time stamps. We observe the sharpness in the VPN feature maps compared to that of I3D for \textit{thumbs down} action which is localized over a small space. For similar actions like \textit{put on headphone} and \textit{flicking hair}, along with salience precision of the VPN feature map, the activations of their corresponding receptive fields show the discriminative power of VPN.

\begin{figure}
\centering
\includegraphics[width=1\linewidth]{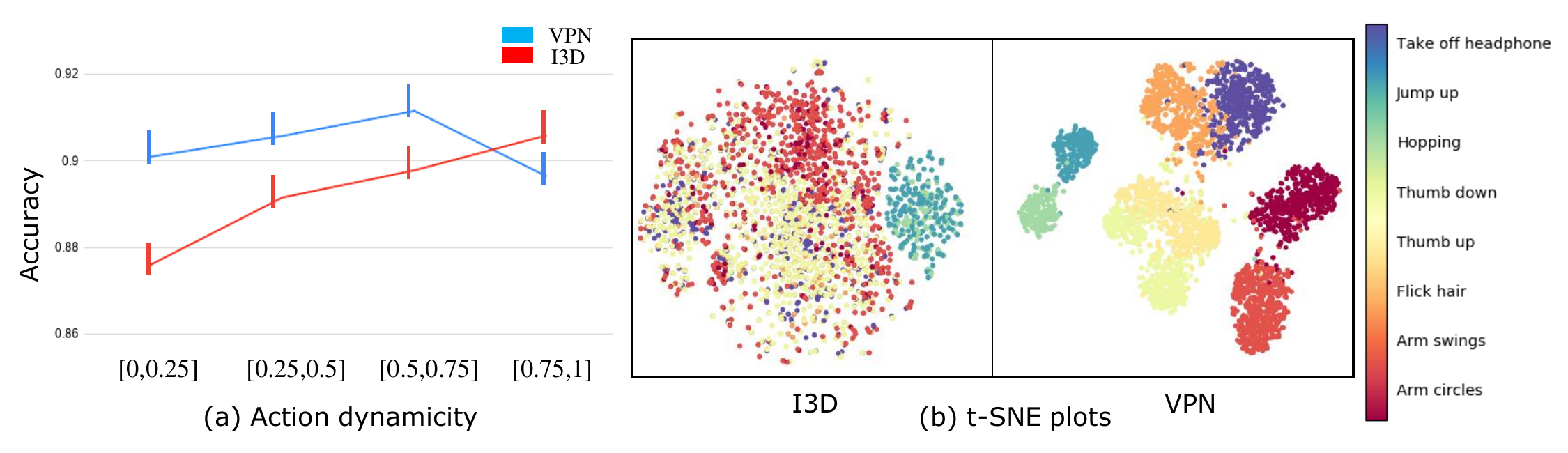}
\caption{ (a) We compare our model against baseline I3D across action dynamicity. Our model significantly improves for most actions. (b) t-SNE plots of feature spaces produced by I3D and VPN for similar appearance actions. }\vspace{-1em}
\label{visualization}
\end{figure}

\noindent In fig.~\ref{visualization}(a), we illustrate the performance of VPN w.r.t. I3D baseline for the dynamicity of an action along the videos. This dynamicity is computed by averaging the Procrustes distance~\cite{procustes} between subsequent 3D poses along the videos. If the average distance is large, it means the poses change a lot in an action. VPN significantly improves for actions with subtle motion like \textit{hush} (+52.7\%), \textit{staple book} (+40.7\%) and \textit{reading} (+36.2\%) which indicates the efficacy of VPN for fine-grained actions. The degradation of the VPN performance for high action dynamicity is negligible(-0.8\%).
 In fig.~\ref{visualization}(b), we show the t-SNE plots of the feature spaces produced by I3D and VPN for some selected actions with similar appearance. It clearly shows the discriminative power of VPN for actions with similar appearance which is a frequent challenge in ADL.

\begin{table}[!ht]
\caption{Results (accuracies in \%) on NTU-60 with cross-subject (CS) and cross-view (CV) settings (at left) and NTU-120 with cross-subject ($CS_1$) and cross-setup ($CS_2$) settings (at right); Att indicates attention mechanism, $\circ$ indicates that the modality has only been used for training, the methods indicated with $^*$ are reproduced on this dataset. 3D ResNeXt-101 is abbreviated as RNX3D101.}
 
\begin{minipage}[b]{0.5\linewidth}
\centering
\scalebox{0.7}{
\begin{tabular}{lccccc}
\hline
Methods & Pose & RGB & Att & CS & CV  \\
\hline\hline
 AGC-LSTM~\cite{skel_gcn_3} & \checkmark & $\times$ & \checkmark & 89.2 & 95.0 \\
 DGNN~\cite{directed_graph} & \checkmark & $\times$ & $\times$ & 89.9 & 96.1 \\
\hline
STA-Hands~\cite{STA-hands} & \checkmark & \checkmark  & \checkmark & 82.5 & 88.6  \\
altered STA-Hands~\cite{Baradel_BMVC} & \checkmark & \checkmark  & \checkmark & 84.8 & 90.6 \\
\small{Glimpse Cloud~\cite{glimpse}} & $\circ$ & \checkmark & \checkmark & 86.6 & 93.2 \\
PEM~\cite{pem} & \checkmark  & \checkmark  & $\times$  & 91.7 & 95.2  \\
Separable STA~\cite{STA_iccv} & \checkmark & \checkmark &\checkmark  & 92.2 & 94.6  \\
P-I3D~\cite{spatial-i3d} & \checkmark & \checkmark & \checkmark & 93 & 95.4 \\
\hline
\textbf{VPN} & \checkmark & \checkmark &\checkmark  & \textbf{93.5} & \textbf{96.2}  \\
\textbf{\small{VPN (RNX3D101)}} & \checkmark & \checkmark &\checkmark  & \textbf{95.5} & \textbf{98.0}  \\
\hline
\end{tabular}
}
\end{minipage}
\begin{minipage}[b]{0.45\linewidth}
\centering
\scalebox{0.7}{
\begin{tabular}{lccccc}
\hline
Methods & Pose & RGB & Att & $CS_1$ & $CS_2$  \\
\hline\hline
ST-LSTM~\cite{st-lstm} &\checkmark  & $\times$  &\checkmark  & 55.7 & 57.9  \\
Two stream Att LSTM~\cite{global-context} &\checkmark  & $\times$  &\checkmark  & 61.2 & 63.3  \\
Multi-Task CNN~\cite{clip_representation} &\checkmark  & $\times$  & $\times$  & 62.2 & 61.8  \\
PEM~\cite{pem} &\checkmark  & $\times$  &\checkmark  & 64.6 & 66.9  \\
2s-AGCN~\cite{2sagcn2019cvpr} &\checkmark  & $\times$  &\checkmark  & 82.9 & 84.9  \\
\hline
Two-streams~\cite{twostream} & $\times$ &\checkmark & $\times$ & 58.5 & 54.8 \\
I3D$^*$~\cite{i3d} & $\times$ & \checkmark & $\times$ & 77.0 & 80.1 \\
\hline
\small{Two-streams + ST-LSTM}~\cite{ntu120} & \checkmark & \checkmark & $\times$ & 61.2 & 63.1 \\
Separable STA$^*$~\cite{STA_iccv} & \checkmark & \checkmark &\checkmark  & 83.8 & 82.5  \\
\hline
\textbf{VPN} & \checkmark & \checkmark &\checkmark  & \textbf{86.3} & \textbf{87.8}  \\
\hline
\end{tabular}
}
\end{minipage}
\label{ntu_accuracy}
\end{table}
\begin{table*}[htp]

\floatsetup{floatrowsep=qquad,captionskip=5pt} \tabcolsep=4pt

\begin{floatrow}
\floatsetup{floatrowsep=qquad,captionskip=5pt} \tabcolsep=4pt
\begin{floatrow}
\ttabbox{\caption{Results on Smarthome dataset with cross-subject (CS) and cross-view ($CV_1$ and $CV_2$) settings (accuracies in \%). 
Att indicates attention mechanism. 
}}{%
\scalebox{0.7}{
\begin{tabular}{lcccccc}
\hline
Methods & Pose & RGB & Att & CS & $CV_1$ & $CV_2$  \\
\hline\hline
DT~\cite{DT} & $\times$ & \checkmark & $\times$ & 41.9 & 20.9  & 23.7 \\
LSTM~\cite{lstm3d} & \checkmark & $\times$ & $\times$ & 42.5 & 13.4 & 17.2 \\
I3D~\cite{i3d} & $\times$ & \checkmark & $\times$ &53.4  & 34.9 & 45.1 \\
I3D+NL~\cite{nonlocal} & $\times$ & \checkmark & \checkmark &  53.6 & 34.3 & 43.9 \\
P-I3D~\cite{spatial-i3d} & \checkmark & \checkmark & \checkmark & 54.2  & 35.1 & 50.3  \\
Separable STA~\cite{STA_iccv} & \checkmark & \checkmark & \checkmark & 54.2  & 35.2 & 50.3  \\
\hline
\textbf{VPN}  & \checkmark & \checkmark &\checkmark  & \textbf{60.8} & \textbf{43.8} & \textbf{53.5}  \\
\hline
\end{tabular}
}
\label{smarthome_accuracy}}
\end{floatrow}
\hspace{3pt}
\begin{floatrow}
  \centering
 \ttabbox{\caption{Results on N-UCLA dataset with cross-view  $V^3_{1,2}$ settings (accuracies in \%); $\overline{Pose}$ indicate its usage only in the training phase.}}{%
\scalebox{0.7}{
\begin{tabular}{lccc}
\hline
Methods & Data & Att & $V^3_{1,2}$  \\
\hline\hline
HPM+TM~\cite{hpm} & Depth & $\times$ & 91.9 \\
Ensemble TS-LSTM~\cite{tslstm} & Pose & $\times$ & 89.2 \\
NKTM~\cite{nktm} & RGB & $\times$ & 85.6 \\
Glimpse Cloud~\cite{glimpse} & RGB+ $\overline{Pose}$ & \checkmark & 90.1 \\
Separable STA~\cite{STA_iccv} & RGB+Pose& \checkmark & 92.4  \\
P-I3D~\cite{spatial-i3d} & RGB+Pose& \checkmark & 93.1  \\
\hline
\textbf{VPN} & RGB+Pose& \checkmark & \textbf{93.5}  \\
\hline
\end{tabular}
}
\label{ucla_accuracy}
}
\end{floatrow}
\end{floatrow}
\vspace{-10 pt}
\end{table*}

\subsection{Comparison with the state-of-the-art}
We compare VPN to the state-of-the-art (SoA) on NTU-60, NTU-120, Smarthome and N-UCLA in table \ref{ntu_accuracy}, \ref{smarthome_accuracy} and~\ref{ucla_accuracy}. VPN outperforms on each of them. In table~\ref{ntu_accuracy} (at left), for input modality RGB+Poses, VPN improves the SoA~\cite{spatial-i3d} by up to 0.8\% on NTU-60 even by using one-third parameters compared to~\cite{spatial-i3d}. The SoA using Poses only~\cite{directed_graph} yields classification accuracy near to VPN for cross-view protocol (with 0.1\% difference) due to their robustness to view changes. However, the lack of appearance information restricts these methods~\cite{directed_graph,skel_gcn_3} to disambiguate actions with similar visual appearance, thus resulting in lower accuracy for cross-subject protocol. We have also tested VPN with 3D ResNeXt-101~\cite{can_spatio-temporal} on NTU-60 dataset. The results in table \ref{ntu_accuracy} show that VPN can be adapted with other existing video backbones.

Compared to the SoA results, the improvement by 3.9\% and 4.9\% (averaging over the protocols) on NTU-120 and Smarthome respectively are significant.   
It is worth noting that VPN improves further the classification of actions with similar appearance as compared to Separable STA~\cite{STA_iccv}. For example, actions like \textit{clapping} (+44.3\%) and \textit{flicking hair} (+19.1\%) are now discriminated with better accuracy. In addition, the superior performance of VPN in cross-view protocol for both NTU-120 and Smarthome implies that it provides better view-adaptive characterization compared to all the prior methods.\\
For N-UCLA which is a small-scale dataset, we pre-train the visual backbone with NTU-60 for a fair comparison with~\cite{glimpse,STA_iccv,spatial-i3d}. We also outperform the SoA~\cite{spatial-i3d} by 0.4\% on this dataset.

\section{Conclusion}
This paper addresses the challenges of ADL classification. We have proposed a novel Video-Pose Network \textbf{VPN} which provides an accurate video-pose embedding. We show that the embedding along with attention network yields a more discriminative feature map for action classification. 
The attention network leverages the topology of the human joints and with the coupler provides precise spatio-temporal attention weights along the video.

 Our recognition model outperforms the state of-the-art results for action classification on 4 public datasets. 
This is a first step towards combining RGB and Pose through an explicit embedding. A future perspective of this work is to exploit this embedding even in case of noisy 3D poses in order to also boost action recognition for internet videos. This embedding could even help to refine these noisy 3D poses in a weakly supervised manner.

\section*{Acknowledgement}
We are grateful to INRIA Sophia Antipolis - Mediterranean "NEF" computation cluster for providing resources and support.
%
%
\bibliographystyle{splncs04}
\bibliography{egbib}
\newpage
\setcounter{section}{0}
\noindent \large{\textbf{Appendix overview}} \\
We provide in section 1 computational details regarding the normalization of Euclidean loss provided in Spatial Embedding of RGB and Pose (section 3.2 (II)). Section 2 provides the details of the baseline with LSTM pose backbone with or without coupler in Table 2 \& 4 from the ablation studies. Section 3 provides the details of the divergence losses used for comparing with Normalized Euclidean loss in Table 3 from ablation studies. Finally, we provide some more insights about VPN in section 4 to illustrate its effectiveness. \\
For convenience, we use the same notation as in the main paper for this supplementary material.

\section{Details on normalization of Euclidean loss}
In equation (4), $\widehat{T_vf_s} = \frac{T_vf_s}{||T_vf_s||_2} = \frac{f_e}{||f_e||_2}$ and $\widehat{T_pz_1} = \frac{T_pz_1}{||T_pz_1||_2} = \frac{P_e}{||P_e||_2}$ are the feature representations projected to the unit hypersphere. Here, we compute the norm $||f_e||_2$ and $||P_e||_2$ using
\begin{equation}
    ||f_e||_2 = \sqrt{\Sigma_if_{e_i}^2 + \epsilon} \hspace{0.3 in}\& \hspace{0.3 in} ||P_e||_2 = \sqrt{\Sigma_iP_{e_i}^2 + \epsilon}
\end{equation}
where $\epsilon$ is a small positive value to prevent dividing zero.

\section{LSTM Pose backbone with or without coupler baselines}
For the LSTM Pose Backbone in Table 2 \& 4, we use a 3-layer stacked LSTM, pre-trained for action classification, as a Pose Backbone by freezing the weights of their cell gates following~\cite{STA_iccv}. The output feature vector $h^*$ is computed by concatenating all the LSTM output features over time. To have a fair comparison with our GCN Pose Backbone, we also introduced residual connections between the original pose input and the LSTM output tensor. However, these residual connections do not improve the action classification accuracy. 

For the experiments in Table 2 to implement the attention network without the coupler, we do not compute  $A_{ST}$. Instead, we multiply the attention weights $inflate(A_S)$ and $inflate(A_T)$ separately with the RGB feature map $f$ in two streams following~\cite{STA_iccv}. Finally, the modulated feature maps from both the streams are concatenated to classify the actions. 

\section{Baselines with KL divergence loss}
In Table 3, we compare different forms of KL divergence loss with normalized euclidean loss for spatial embedding of RGB and 3D poses.
The KL-divergence losses $D_{KL}(f_e||P_e)$ and $D_{KL}(P_e||f_e)$ for n samples are computed by
\begin{align}
    D_{KL}(f_e||P_e) = \sum\limits_{i=1}^{n}f_e^ilog(\frac{f_e^i}{P_e^i}) \\
    D_{KL}(P_e||f_e) = \sum\limits_{i=1}^{n}P_e^ilog(\frac{P_e^i}{f_e^i})
\end{align}
where $f_e^i$ and $P_e^i$ are visual and pose embedding of the $i^{th}$ input sample.\\
Finally, the bi-directional KL-divergence loss is given by $D_{KL}(f_e||P_e) + D_{KL}(P_e||f_e)$.

\section{Detailed qualitative analysis of VPN}
In this section, we provide illustrations to show the impact of each VPN components in section 4.1, superiority of VPN compared to other representative baselines in section 4.2, and some result visualization to highlight the solved and remaining challenges in ADL.
\subsection{Illustration to show the impact of VPN components}
\begin{figure*}
\centering
\includegraphics[width=1\linewidth]{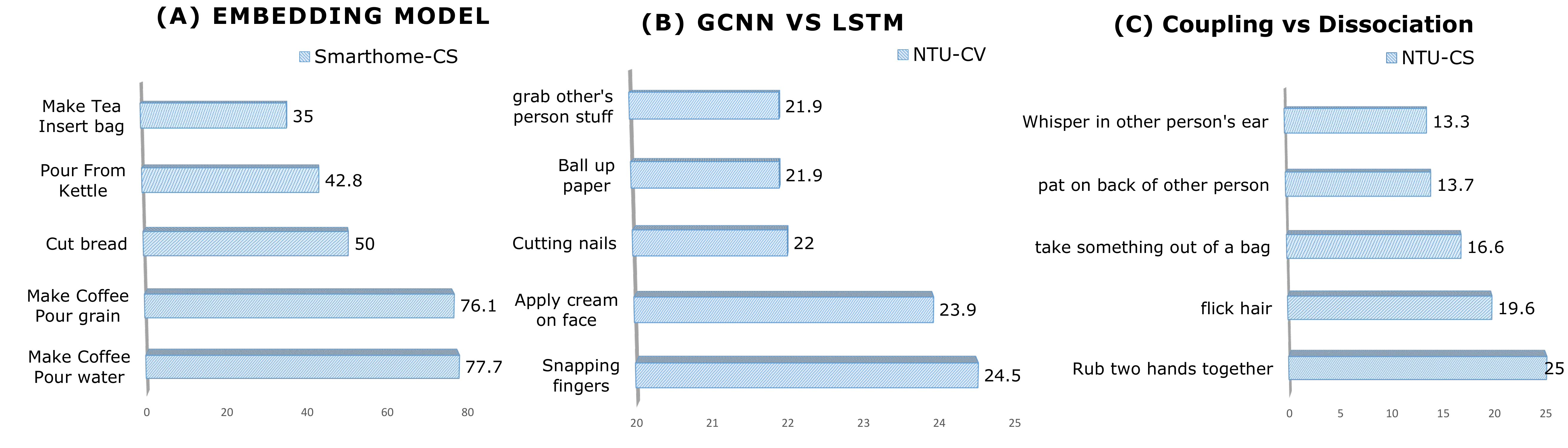}
\caption{Graphs illustrating the superiority of each component of VPN compared to their counterparts (without the respective components). We present the Top-5 per class improvement for (a) VPN with embedding vs without embedding (only Spaital Attention), (b) VPN with GCN vs LSTM Pose Backbone, and (c) attention in VPN with vs without spatio-temporal coupler.}\vspace{-1em}
\label{module_graph}
\end{figure*}

In fig.~\ref{module_graph}, we illustrate a set of graphs showing the top-5 improvement of action classification accuracy using different components of VPN compared to I3D baseline. As discussed in the ablation studies of the primary paper, each component in VPN is critical for good performance on ADL recognition.
\begin{itemize}
    \item The spatial embedding provides an accurate alignment of the RGB images and the 3D poses. As a result, the recognition performance of the fine-grained actions improves compared to its counterpart without embedding (see fig.~\ref{module_graph} (a)).
    \item The GCN pose backbone of the attention network, not only provides a strategy to globally optimize the recognition model but also takes the human joint configuration into account for computing the attention weights. This further boosts the action classification performance (see fig.~\ref{module_graph} (b)).
    \item The spatio-temporal coupler of the attention network provides discriminative spatio-temporal attention weights which enables the recognition model to better disambiguate the actions with similar appearance (see fig.~\ref{module_graph} (c)).
\end{itemize}

\subsection{Illustration to show the superiority of VPN}
\begin{figure*}
\centering
\includegraphics[width=1\linewidth]{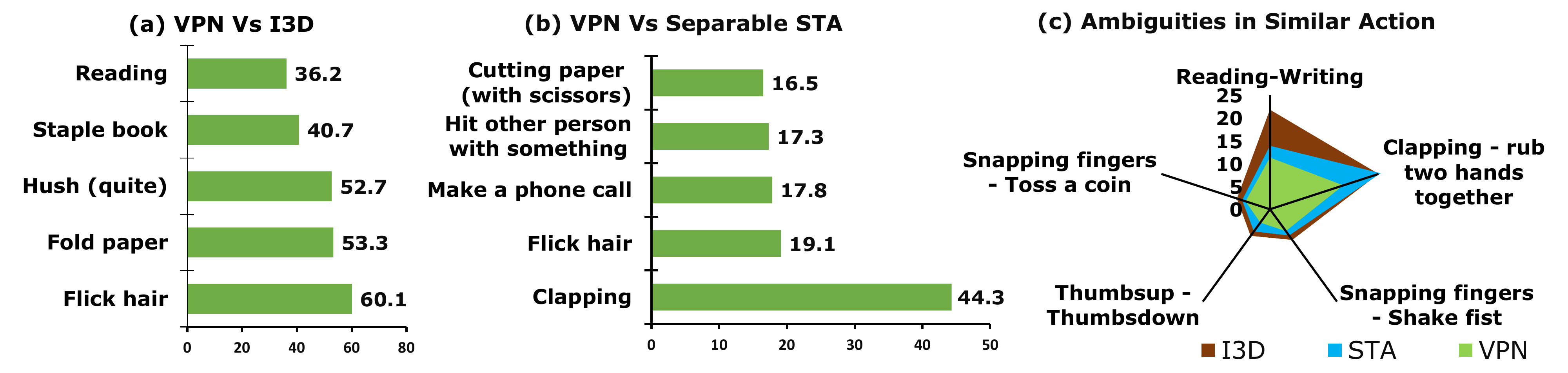}
\caption{Graphs illustrating the superiority of VPN compared to the state-of-the-art methods. We present the Top-5 per class improvement for VPN over (a) I3D baseline and (b) Separable STA. In (c), we present a radar for the average mis-classification score of few action-pairs: lower scores indicate lesser ambiguities between the action-pairs. }\vspace{-1em}
\label{soa_graph}
\end{figure*}

We illustrate in fig.~\ref{soa_graph}, the top-5 per-class classification improvement compared to baseline I3D~\cite{i3d} and to an attention mechanism (Separable STA~\cite{STA_iccv}) from the state-of-the-art, utilizing 3D poses. The significant accuracy improvements for actions with subtle motion like \textit{hush} (+52.7\%), \textit{staple book} (+40.7\%) and \textit{reading} (+36.2\%) as depicted in fig.~\ref{soa_graph}{ {(a)}} illustrate the efficacy of VPN for fine-grained actions. It is worth noting that VPN improves further the classification of actions possessing similar appearance as compared to separable STA in fig.~\ref{soa_graph}{ {(b)}}. For example, actions like \textit{clapping} (+44.3\%) and \textit{flicking hair} (+19.1\%) are now discriminated with better accuracy. Further, in fig.~\ref{soa_graph}{ {(c)}} we present a radar for the average mis-classification score of few action-pairs. The smaller area under the curve for VPN compared to I3D baseline and Separable STA shows that it is able to better disambiguate the action-pairs even with low inter-class variation.

\subsection{Result visualization}
\begin{figure*}
\centering
\includegraphics[width=1\linewidth]{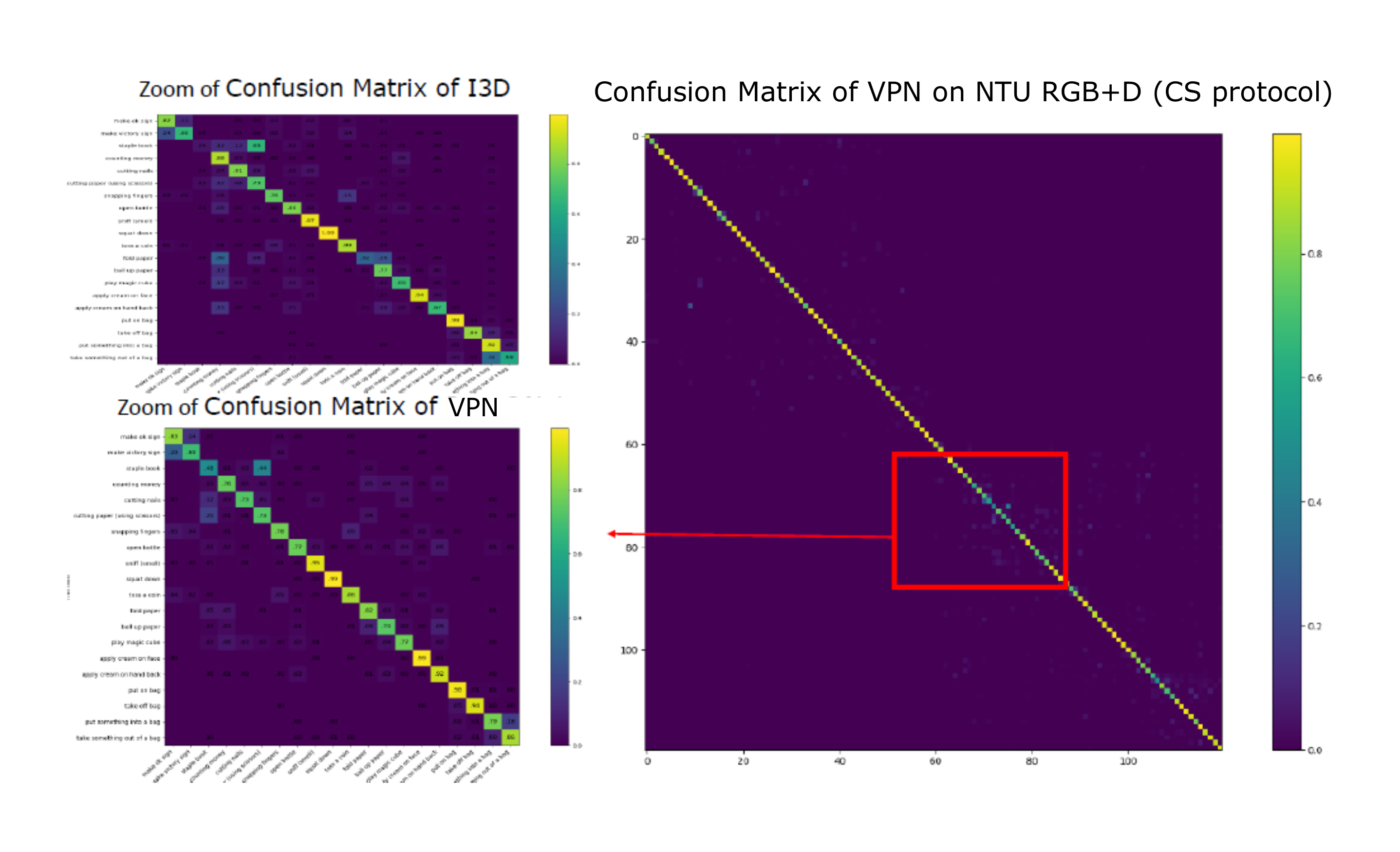}
\caption{Confusion matrix of VPN on NTU RGB+D (CS Protocol) on the right. Zoom of the red bounding box on the left along with the corresponding confusion matrix of I3D. }
\label{cnf_ntu_cs}
\end{figure*}

In this section, we provide the confusion matrix for action classification on NTU RGB+D 120 and Toyota Smarthome using VPN.  
In fig~\ref{cnf_ntu_cs}, we present the confusion matrix of VPN on NTU RGB+D (on right) and a zoom of it around the red bounding box (on left). We also present the corresponding zoom of the confusion matrix of I3D. We are particularly interested in the mis-classifications performed by VPN and thus, we zoom into the region with relatively low classification accuracy. We observe that actions like \textit{staple book} and \textit{taking something out of bag} were confused with \textit{cutting papers} and \textit{put something into  a bag} respectively when classified with I3D. However, with VPN these actions with similar motion are now better discriminated, improving their classification accuracy by approximately  42\% and 27\% respectively. 

\begin{figure*}
\centering
\includegraphics[width=1\linewidth]{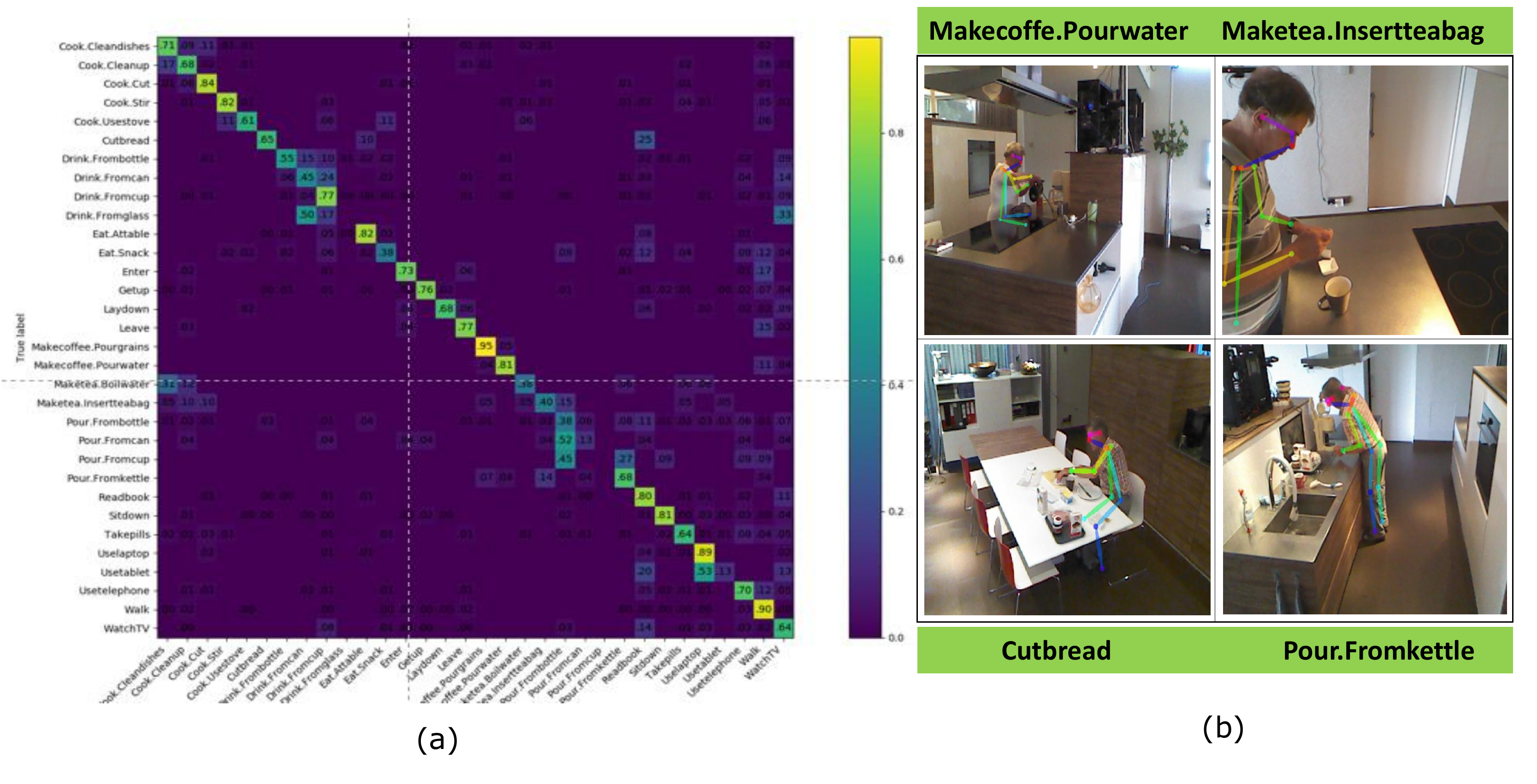}
\caption{(a) Confusion matrix of VPN on Toyota Smarthome (CS protocol) (b) Illustration of poses for activities  mis-classified with I3D but correctly classified with VPN. }
\label{cnf_smarthome_cs}
\end{figure*}

\begin{figure*}
\centering
\includegraphics[width=1\linewidth]{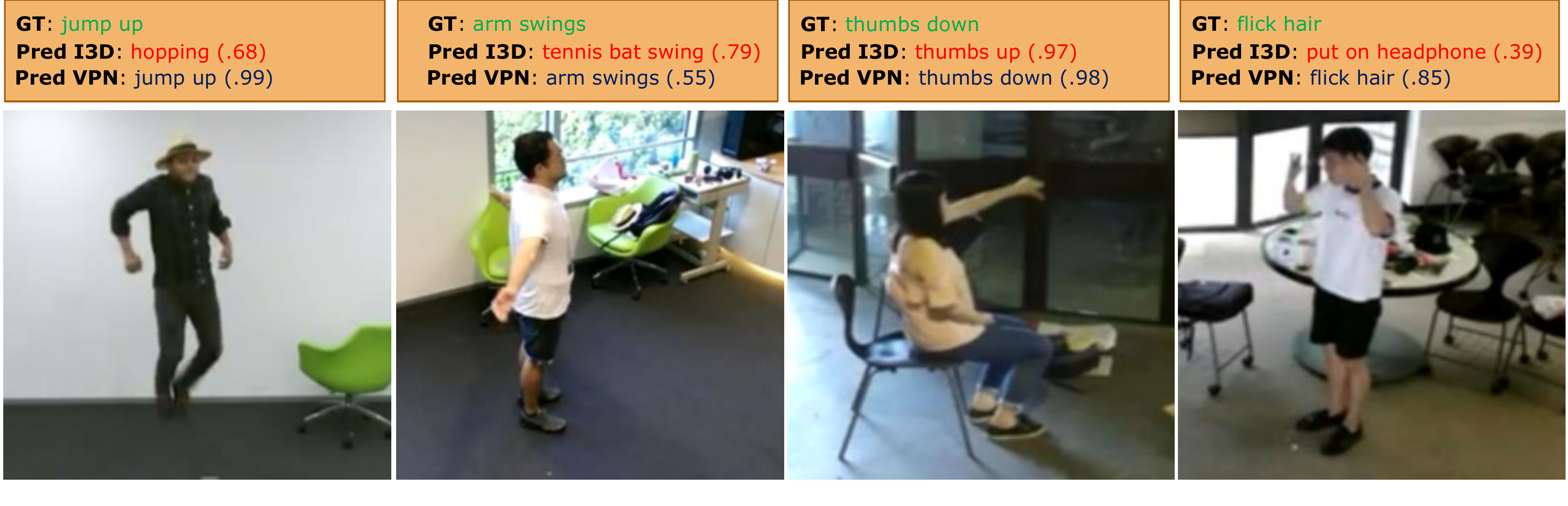}
\caption{Visual results from NTU RGB+D 120 where VPN outperforms I3D.}
\label{visual_results}
\end{figure*}

\begin{figure*}
\centering
\includegraphics[width=1\linewidth]{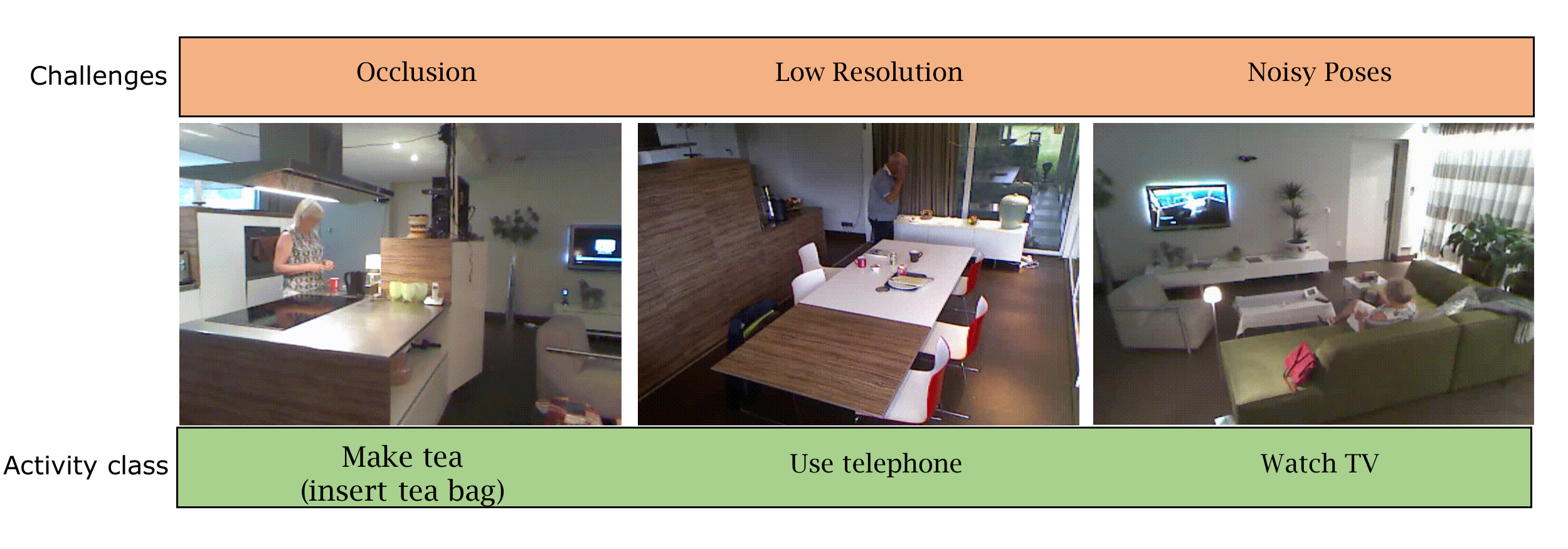}
\caption{Illustration of the remaining challenges in Toyota Smarthome with images from activities (indicated below) and their corresponding challenges (indicated on the top)}
\label{challenges}
\end{figure*}

Similarly, in fig.~\ref{cnf_smarthome_cs} (a), we present the confusion matrix of VPN on Toyota Smarthome dataset. In fig.~\ref{cnf_smarthome_cs} (b), we show the poses for some images belonging to action videos mis-classified by I3D. Thanks to the high quality 3D poses for these videos, now VPN can correctly classify these actions taking the human topology of the 3D poses into account. We provide some visual results in fig.~\ref{visual_results} where VPN outperforms I3D baseline.
We notice that actions like \textit{Drink from glass} are not recognized due to extremely low number of training samples. We further notice that actions like \textit{using tablet} are recognized with low accuracy of 13\% and largely confused with \textit{using laptop}. However, I3D completely mis-classifies the action \textit{using tablet}. 
We also observe that still few action classes are recognized with extremely low classification accuracy. We infer that these poor classification results on certain videos are due to occlusion, low resolution of the actions and low quality poses as illustrated in fig ~\ref{challenges}. 

\end{document}